\pdfoutput=1

\documentclass[11pt]{article}

\usepackage[final]{acl}

\usepackage{times}
\usepackage{latexsym}

\usepackage[T1]{fontenc}

\usepackage[utf8]{inputenc}

\usepackage{microtype}

\usepackage{inconsolata}

\usepackage{graphicx}

\usepackage{multirow}
\usepackage{amssymb}
\usepackage{booktabs}
\usepackage{tabularx}
\usepackage{caption}
\usepackage{hyperref}

\title{Label Words as Local Task Vectors in In-Context Learning}
\author{
	\textbf{Bowen Zheng\textsuperscript{1,*}},
	\textbf{Ming Ma\textsuperscript{1,2,*}},
	\textbf{Zhongqiao Lin\textsuperscript{1}},
	\textbf{Tianming Yang\textsuperscript{1,\dag}},
	\\
	\\
	\textsuperscript{1}Institute of Neuroscience, Key Laboratory of 
	\\
	Brain Cognition and Brain-inspired Intelligence Technology, 
	\\
	Center for Excellence in Brain Science and Intelligence Technology, 
	\\
	Chinese Academy of Sciences, Shanghai, China,
	\\
	\textsuperscript{2}University of Chinese Academy of Sciences
    \\
    \small{\textbf{Correspondence:} \href{mailto:email@domain}{\{zhengbw, mam2022, zqlin, tyang\}@ion.ac.cn}} \\
    \small{*These authors contributed equally to this work.}
	}

\begin{document}
\maketitle
\begin{abstract}
Large Language Models (LLMs) have demonstrated remarkable abilities, one of the most important being in-context learning (ICL). With ICL, LLMs can derive the underlying rule from a few demonstrations and provide answers that comply with the rule. Previous work hypothesized that the network creates a task vector in specific positions during ICL. The task vector can be computed by averaging across the dataset. It conveys the overall task information and can thus be considered global. Patching the global task vector allows LLMs to achieve zero-shot performance with dummy inputs comparable to few-shot learning. However, we find that such a global task vector does not exist in all tasks, especially in tasks that rely on rules that can only be inferred from multiple demonstrations, such as categorization tasks. Instead, the information provided by each demonstration is first transmitted to its answer position and forms a local task vector associated with the demonstration. In some tasks but not in categorization tasks, all demonstrations' local task vectors converge in later layers, forming the global task vector.  We further show that local task vectors encode a high-level abstraction of rules extracted from the demonstrations. Our study provides novel insights into the mechanism underlying ICL in LLMs, demonstrating how ICL may be achieved through an information aggregation mechanism.
\end{abstract}
\label{sec:abs}

\begin{figure*}[ht]
\centering
\includegraphics[width=\linewidth]{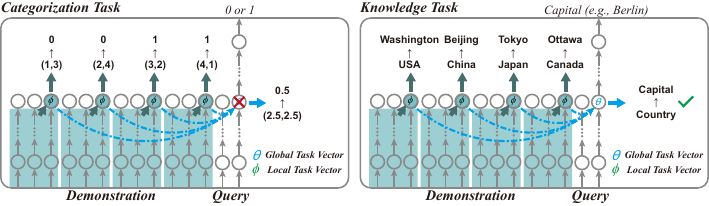}
\caption{Illustration of local and global task vectors. Left: In a categorization task, local task vectors contain information associated with each demonstration, but they do not converge and form a global task vector. Right: In a knowledge task, local task vectors aggregate into a coherent global task vector that aligns with the LLM’s prior knowledge, enabling effective task representation.}
\label{fig:Main Graph}
\end{figure*}

\section{Introduction}
\label{sec:intro}
The advent of Large Language Models (LLMs) has enabled machines to understand and generate human-like text with unprecedented accuracy. One of the most remarkable abilities of LLMs is in-context learning (ICL), where the model can abstract the underlying rule defined by a few demonstrations and provide answers that comply with that rule \cite{brown2020language, liu2023pre, dong2023survey}. This capability has garnered significant attention from the research community, as it demonstrates the flexibility of LLMs to adapt to new tasks without extensive training, which is a signature of human cognition \cite{binz2023using}. Unlike prompt fine-tuning \cite{lester2021power} or chain-of-thought prompting \cite{wei2022chain}, ICL simply relies on several demonstrations that share the same structure as the question. 

Previous mechanistic work suggests that task information in ICL can be represented as compact vectors localized at specific token positions and layers in the model.~\cite{hendel2023incontext, wang2023label, pmlr-v235-liu24bx, li2024context}.
Specifically, the task vector is computed by averaging the embedding of the last token at a particular layer across samples from the same task dataset. Researchers have shown that by transferring the task vector to the corresponding positions with zero-shot dummy inputs, LLMs can achieve performance similar to few-shot learning. This task vector is therefore deemed to carry a full abstraction of the task. We term it the global task vector to distinguish it from the local task vector defined below.

While patching the global task vector works as expected in some tasks (e.g., knowledge tasks), we find that it does not work well in other tasks, particularly those in which the rule can only be inferred from multiple demonstrations (e.g. learning an arbitrary categorization boundary).
This is consistent with recent large-scale evidence that complex ICL tasks may rely on multiple subtask-specific vectors rather than a single averaged task vector~\cite{tikhonov2025one}.
Instead, we find that patching the tokens at the \textit{answer} positions of each demonstration to the corresponding positions of a dummy input sequence leads to higher accuracy in these situations. Inspired by the previous study that reveals the importance of label words in the information flow of ICL \cite{wang2023label}, we propose that these label words serve as local task vectors that carry the task rule information in a distributed manner. In addition, we show that even in tasks where a global task vector exists, the local task vectors appear in earlier layers than the global task vector. The information contained in the local task vectors converges and forms the global task vectors in later layers.

Taken together, these results point to a simple picture of ICL: each demonstration first writes task-relevant information into the hidden state at its \textit{answer} token, forming a \emph{local task vector}, and the model answers the query by aggregating information from these local vectors.
This picture also helps explain when global task vectors succeed: in knowledge tasks, local task vectors across demonstrations tend to become increasingly aligned in later layers, making an averaged \emph{global} task vector effective, whereas in categorization tasks they can remain heterogeneous and complementary, so averaging may discard essential information.


\section{Preliminaries}
\subsection{In-context Learning}
We use $T$ to denote a decoder-only transformer LLM, $S$ to denote the set of demonstrations used as the inputs to the LLM, and $x$ to denote the query that needs to be answered. We use $T([S, x])$ to denote the output of LLM on the concatenation of $S$ and $x$. For clarity, we define $S=[Q_0 I_0 A_0 D_0, Q_1 I_1 A_1 D_1, ..., Q_{J} I_{J} A_{J} D_{J}]$ and $x=[Q_{J+1} I_{J+1}]$, where $Q_j$, $I_j$, $A_j$, and $D_j$ are the tokens for the \textit{query}, the \textit{is}, the \textit{answer} and the \textit{dot} of demonstration ${j\;(j=1,...,J)}$. For example, “cat is 0. monkeys is 1. dog is”, each demonstration can be divided into four parts: $Q$, $I$, $A$, and $D$.

\subsection{Global Task Vector}
We follow the definition in \cite{hendel2023incontext}. Assume that ICL operates within a hypothesis space. This mode of operation can be defined as a learning algorithm (denoted $\mathcal{A}$). 
$\mathcal{A}$ maps $S$ to \textit{the global task vector} $\theta$. 
Next, the LLM maps the query $x$ to the output through rule application (denoted by $f$), based on $\theta \equiv \mathcal{A}(S)$, without direct dependence on $S$. Here, the global task vector $\theta$ is at position $I_{J+1}$. 
In our experiments, we consider different versions of $\theta$ and $\phi_j$, including those extracted from a single trial and those averaged across samples. We refer to these as \textit{No Avg} and \textit{Trial Avg}, respectively; see Section \ref{sec:eval_details} and Table \ref{tab:main_task_acc} for formal definitions.
Consider the mapping from a set of demonstrations and a query to the predicted output:
\begin{equation}
    T([S, x]) = f(x; \mathcal{A}(S))
\end{equation}
ICL can be viewed as operating on the following hypothesis class: $\mathcal{H} = \{f(\cdot; \theta) \mid \theta\}$. 

\subsection{Saliency Score}
We use the saliency score \cite{michel2019sixteen, simonyan2013deep, wang2023label} to identify the level of attention that the LLM gives to each token in the input. The saliency score is defined as 
\begin{equation}
	{I_l} = \left| {\sum\limits_h {{A_{h,l}}}  \odot \frac{{\partial {\mathcal L}(x)}}{{\partial {A_{h,l}}}}} \right|,
\end{equation}
Here, $A_{h,l}$ is the value of the attention matrix of the $h$-th attention head in the $l$-th layer, $x$ is the input, and $\mathcal{L}(x)$ is the loss function of the task.

\begin{figure}[hb]
    \centering
    \includegraphics[width=0.7\linewidth]{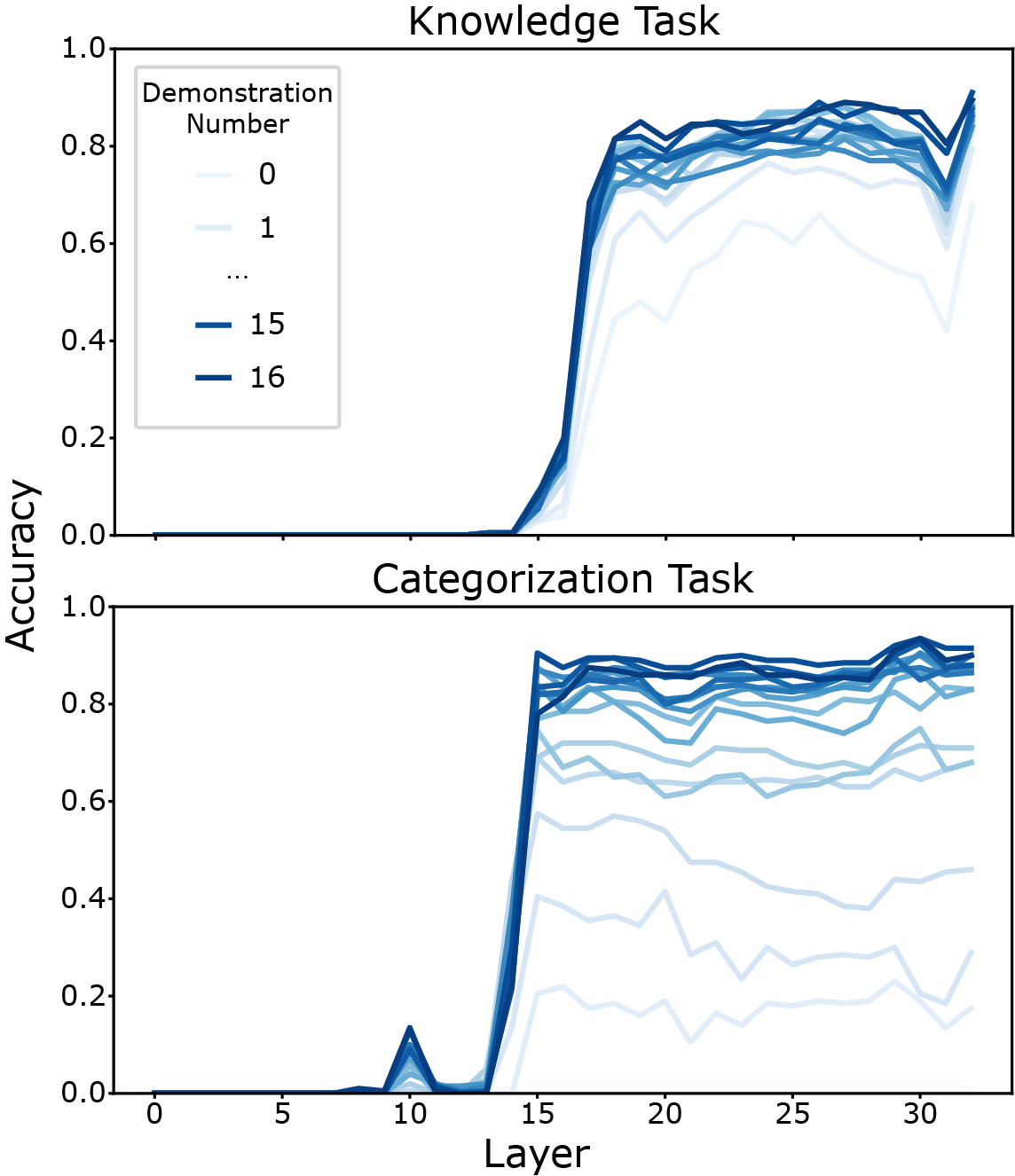}
    \caption{Accuracy increases sharply in the middle layers. The shades of blue indicate demonstration numbers. Top: knowledge task; Bottom: categorization task.} 
\label{fig:acc_by_layers}
\end{figure}

\begin{figure*}[ht]
    \centering
    \includegraphics[width=0.95\linewidth]{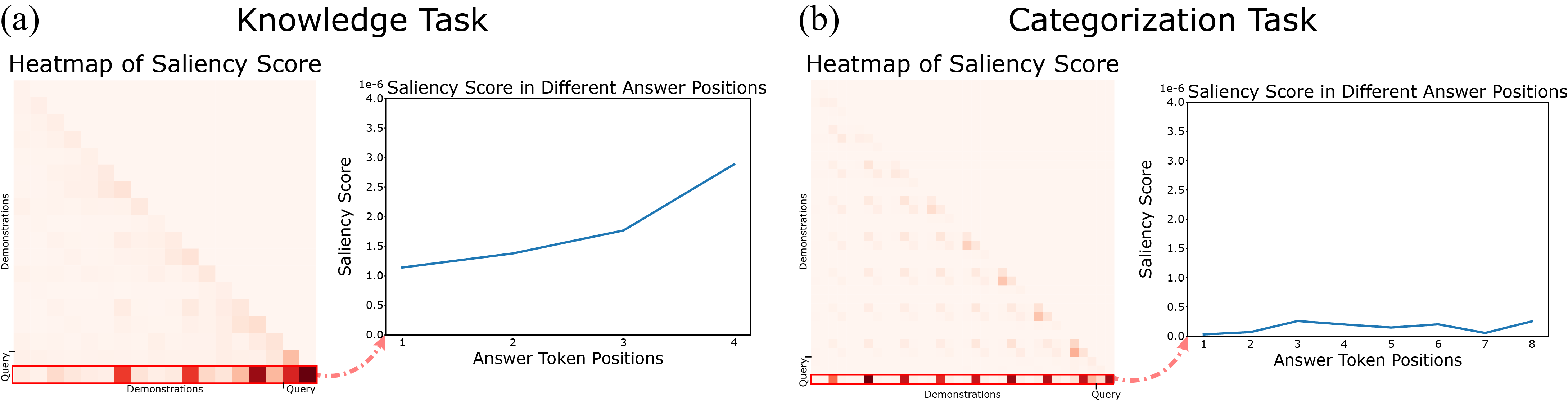}
    \caption{Saliency scores in layer 14, shown as heatmaps. The color at each location shows the saliency score between the positions in the respective row and column. The last row is where the highest scores are observed at the demonstrations' \textit{answer} position, which is plotted on the right as a function of the demonstration index. (a) Knowledge task; (b) Categorization task.}
\label{fig:saliency}
\end{figure*}

\subsection{dPCA}
Following \cite{kobak2016demixed}, we use demixed PCA (dPCA) to isolate the subspace of representations associated with a specific factor. In Sec.~\ref{sec:ans_pos_as_local} we use dPCA to identify the subspace predictive of the demonstration's string length and remove it by projecting answer-position states to the null space of that subspace.
We find a compressing matrix $D$ and a decompressing matrix $F$ by minimizing:

\begin{equation}
    L_{dPCA}= ||{X_L}-{FDX}||^2
\end{equation}
where $X$ is the $h \times{ n }$ activity matrix, where $h$ is the dimension of the hidden state, and $X_L$ is a label average matrix. $X_L$ has the same size as $X$, but all rows are replaced by the mean activity vector with the same label.

We then project the local task vectors onto the null space of the informative subspace: 
\begin{equation}
    {B} = Null({D}) 
\end{equation}
\begin{equation}
    \tilde{{X}} = {B}({B}^T{B})^{-1}{B}^T{X}
\end{equation}
where $D \in \mathbb{R}^{d \times h}$ is the compression matrix, $Null(\,\cdot\,)$ is the null space computation so that ${D}\,\cdot\,Null({D})=0$.  $\tilde{X}$ is the manipulated activity matrix. The mean is subtracted before the operation and then added back when we put the matrix back into the model.

\section{Tasks}
We analyze two representative tasks in depth, and evaluate generality on additional tasks listed in Table~\ref{tab:task_list}. Throughout the paper, we distinguish \textbf{knowledge} tasks, where the input--output mapping is largely consistent with associations likely learned during pretraining (e.g., Country$\to$Capital), from \textbf{rule-induction (categorization)} tasks, where the mapping is defined by the demonstrations in the prompt and must be inferred by aggregating multiple examples (e.g., length thresholding, 2-D boundaries).

The first task is a country-capital knowledge task \cite{hendel2023incontext, wang2023label}. The model should give the capital of a given country. For example, the prompt “China is Beijing. Japan is Tokyo. Germany is” should lead to the answer “Berlin”. In this setup, the final \textit{is} position (after “Germany”) corresponds to the \textbf{global task vector} location, while each \textit{answer} position (e.g., “Beijing”, “Tokyo”) serves as a \textbf{local task vector}. 

Similarly, the second task is a two-alternative categorization task. In this task, the final \textit{is} token before the model prediction is the global task vector position, while each number token (e.g., “0”, “1”) in the demonstrations serves as a local task vector. 

Details of these roles are analyzed in the following sections. In the second task, the model is given a string of random characters, with the length varying between 1 and 10. The model should answer 0 when the string is shorter than 6 characters and 1 otherwise. An example question is “cat is 0. monkeys is 1. dog is”. The number of demonstrations ranges from 1 to 16. We show that models are capable of solving these problems (Fig.~\ref{fig:acc_by_layers}). Consistent with previous work \cite{hendel2023incontext}, the accuracy increases sharply in the middle layers. 

Table \ref{tab:all_task_acc} includes results for more knowledge tasks and categorization tasks.
Unless otherwise noted, we use 4 demonstrations for knowledge tasks and 8 demonstrations for categorization tasks, chosen to yield comparable baseline ICL accuracy and avoid ceiling effects. 

Full details on models, tokenization, and evaluation protocols are deferred to Appendix~\ref{sec:eval_details}.

\begin{figure}[htb]
    \centering
    \includegraphics[width=0.8\linewidth]{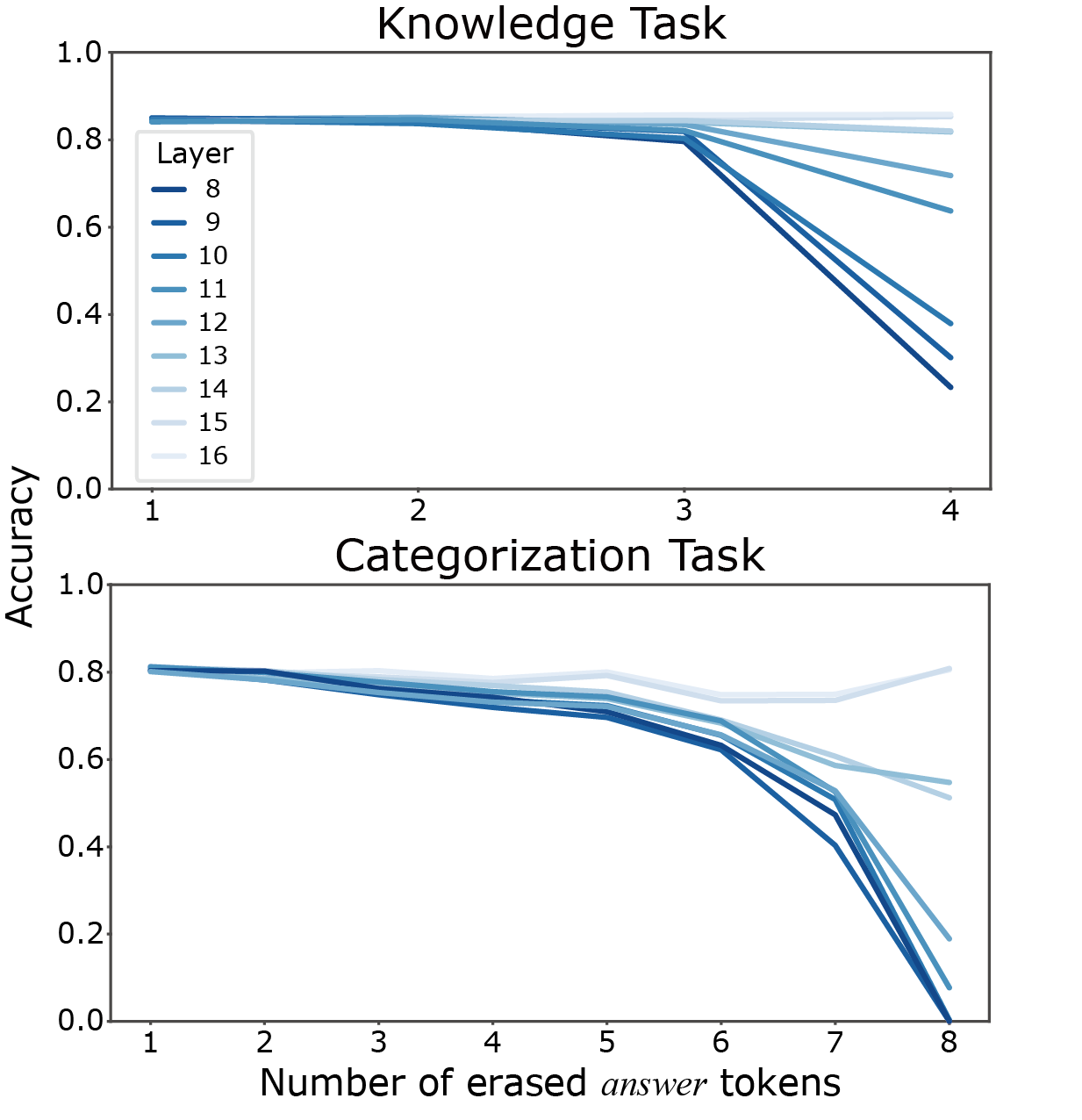}
    \caption{Model accuracy of ablated models. The shades of blue indicate the index of layers. Top: knowledge task; Bottom: categorization task.}
    \label{fig:lesioning}
\end{figure}

\section{Local Task Vectors}\label{headings}
\subsection{The Importance of Answer Positions}
Previous research \cite{hendel2023incontext, wang2023label} shows that the information of each demonstration converges at its respective \textit{answer} position and is important in in-context learning. We further verify this result in the two representative tasks. Using layer 14 as a representative layer, the saliency scores in both tasks are the highest between the \texttt{answer}-position tokens of the demonstrations and the \texttt{is}-position token of the final query, suggesting that the task-related information from each demonstration is transmitted via the corresponding \textit{answer} position's token to the token at the \textit{is} position of the query to form the output (Fig.~\ref{fig:saliency}a). The \textit{is} position token of the query is the global task vector.

In addition, we found that the saliency score is higher for later demonstrations in the knowledge task, suggesting further information aggregation to the \textit{answer} positions of later demonstrations. However, such a trend is not found in the categorization task (Fig.~\ref{fig:saliency}b). Lesion experiments on the answer tokens also reveal the difference between the two tasks. While setting the tokens at the \textbf{answer} positions to zero or random values does not harm the model's performance unless all answer tokens are erased, the same manipulation leads to a gradual decrease in performance in the categorization task (Fig.~\ref{fig:lesioning}). This difference suggests that the ICL mechanism may differ in the two tasks.

\begin{figure}[ht]
\centering
\includegraphics[width=0.9\linewidth]{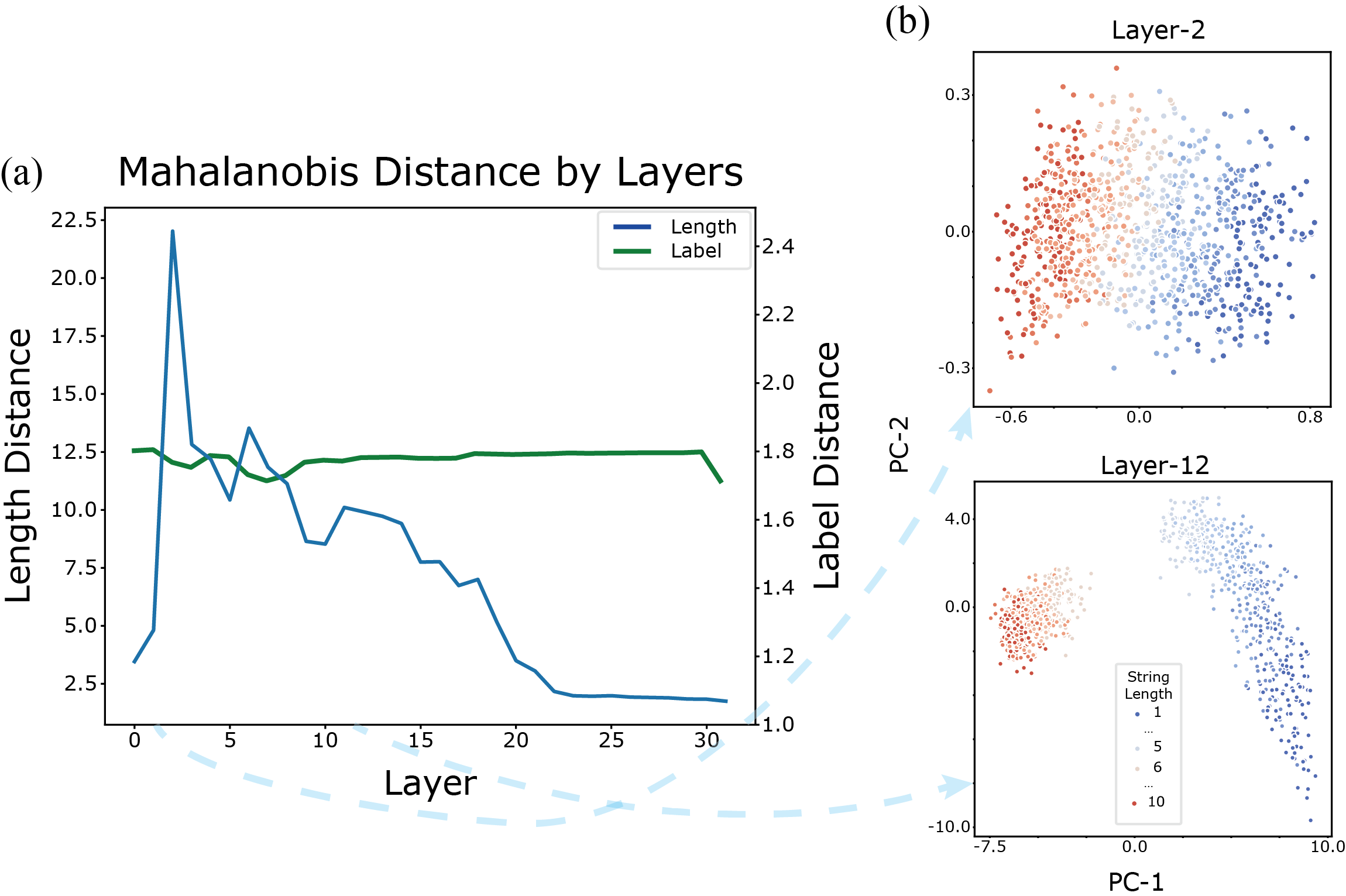}
\caption{The encoding of question and answer information in \textit{answer} positions. (a) Plotted is the Mahalanobis distance of the clusters defined by the demonstrations' string's length (blue) and by the corresponding answer (green) in the space defined by the two largest PCA components of the \textit{answer} positions. Peaking at layer 2, the Mahalanobis distance for the string length gradually decreases across layers. (b) Two example layers' task spaces defined by the first two largest PCA components. Blue and red indicate answers \textit{0} and \textit{1}, and the color gradient indicates string length. Notice that the segregation between the two answers (blue and red color) is maintained across the layers, while the dots with the same color but different gradients are more mixed in later layers (e.g. layer 12) than in earlier layers (e.g. layer 2).}
\label{fig:pca}
\end{figure}

\subsection{Answer Positions As Local Task Vectors}
\label{sec:ans_pos_as_local}
To verify that the information from each demonstration converges at its \textit{answer} position, we use the principal component analysis (PCA) to explore the information encoded at the \textit{answer} positions (Fig.~\ref{fig:pca}). 
In the categorization task, each demonstration provides a string and its associated category label.
For each layer, we collect the hidden states at the demonstrations' \textit{answer} positions from 1000 samples and perform PCA on these vectors (Fig.~\ref{fig:pca}).
To quantify which factor is encoded, we group the answer-position vectors (i) by the string length (1--10) and (ii) by the label token (0/1), and compute the \textit{average pairwise} Mahalanobis distance between group centroids in the space spanned by the top two PCs.

The results suggest that the clusters formed by strings with different lengths are well segregated, with the distance peaking at layer 2 and gradually decreasing. In comparison, the category label is encoded at the \textit{answer} positions stably across layers.

\begin{figure*}[ht]
    \centering
    \includegraphics[width=\linewidth]{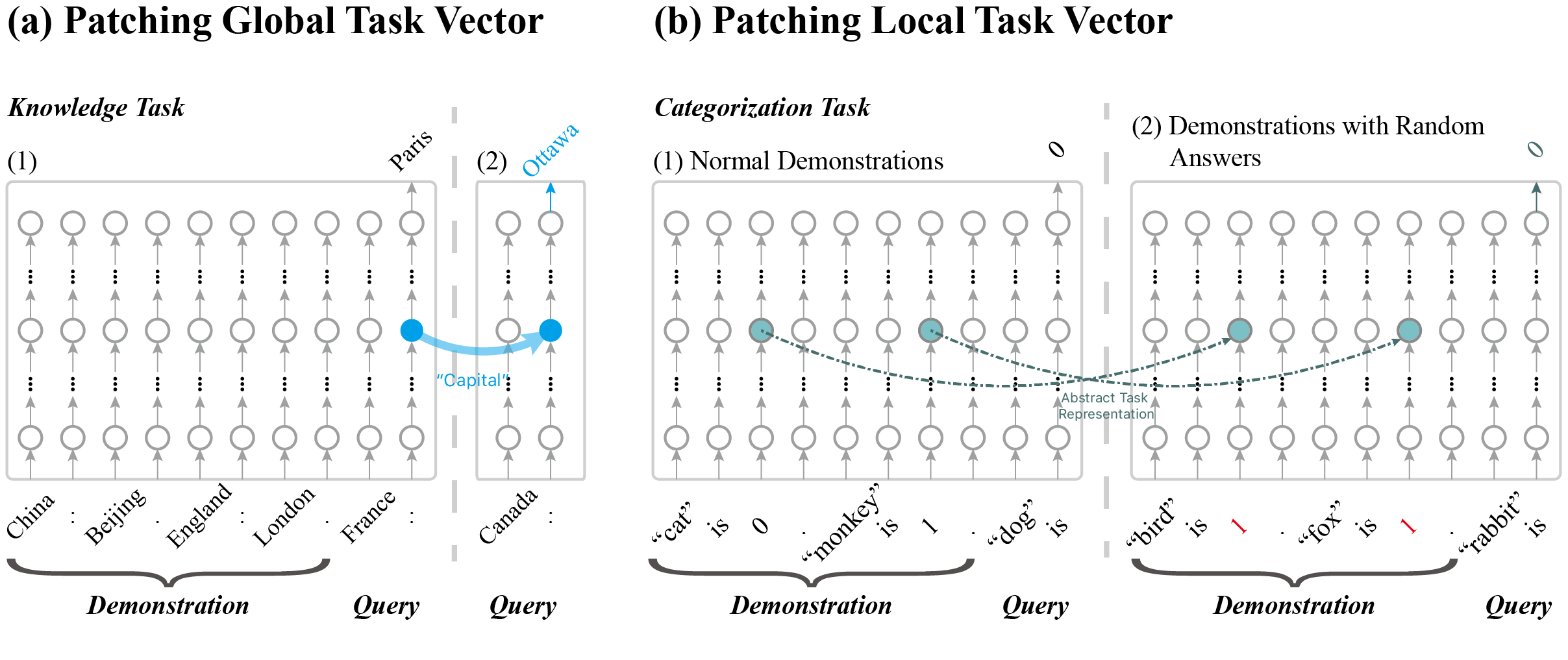}
    \caption{Patching experiments. (a) Patching global task vector. The global task vector, which is the token at the \textit{is} position of the final query, in a network receiving normal demonstrations is copied to a network performing zero-shot inference. (b) Patching distributed local task vectors. The local task vectors, which are tokens at the \textit{answer} position of each demonstration in an LLM receiving normal demonstrations are copied to an LLM receiving the same number of demonstrations but with random \textit{answers}.}
\label{fig:patch flow}
\end{figure*}

To confirm that the string length information contained at the \textit{answer} positions is critical for ICL, we carry out an experiment in which we selectively remove the string length information from the local task vectors. For each \textit{answer}-position token, we first find its subspace that contains the corresponding demonstration's string length information with dPCA and then project the token onto the null space of the found subspace. We set the dimensionality of the discarded subspace to $d=10$. This value strikes a balance between effectively isolating the string-length feature in early layers and preserving task-relevant abstract information in later layers. This procedure erases all information from the subspace that contains the string length information while keeping the rest of the information intact.

\begin{figure}[ht]
    \centering
    \includegraphics[width=0.9\linewidth]{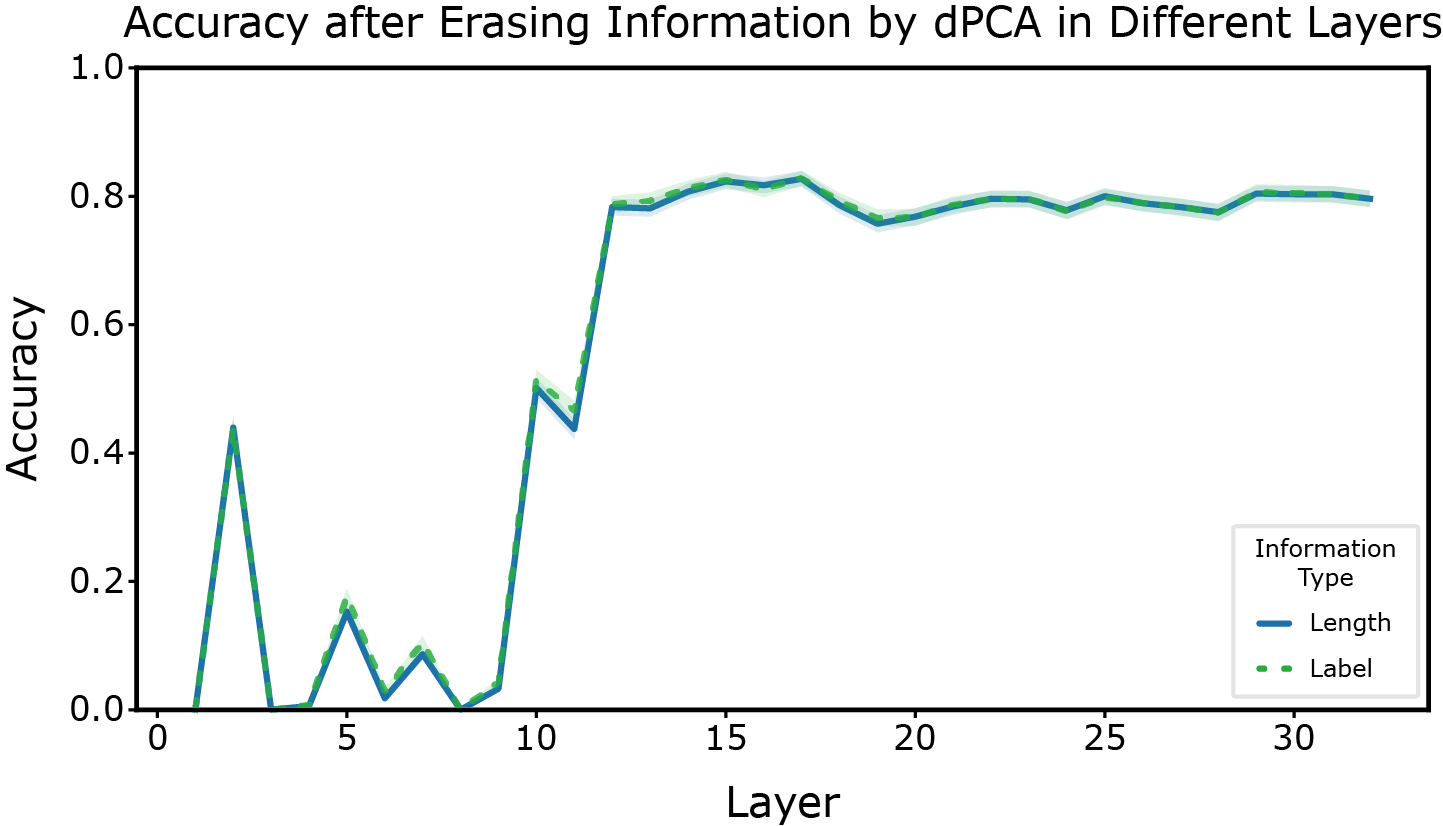}
    \caption{The model's accuracy after string length information in the local task vector is removed in the dPCA space. Note the ablation is only effective in early layers.}    
    \label{fig:dPCA_erase}
\end{figure}

We apply this selective information removal at each layer (Fig.~\ref{fig:dPCA_erase}). Interestingly, the procedure leads to poor performance when applied in the early layers, where the string length is best encoded in the PCA space. However, starting from the middle layers, around layer 13, the ablation results in only a minimal decrease in performance. Nevertheless, in the following sections, we demonstrate that these layers encode sufficient task-relevant information for the model to perform the task effectively. These results suggest that in the later layers, the task information encoded in the answer tokens is a high-level abstraction of \textit{query}-\textit{answer} information that is not affected by selective information removal.

These results suggest that the \textit{answer} position tokens contain task-related information provided by each demonstration and are critical for ICL. As they are local to each demonstration, we term them local task vectors.

\section{Patching Local vs. Global Task Vectors}
\label{sec:patching_vector}
If the local task vectors contain sufficient information for solving the task, copying them to a dummy sequence should yield reasonable performance. Following the previous work \cite{hendel2023incontext}, we patch the local task vectors to a dummy sequence with those from a normal sequence. Here, the dummy input contains random labels and its demonstrations are different from the original sequence to prevent information leakage. More specifically, given an original sequence \(S\) and a dummy sequence \(D\), we patch \(S_a^{j}\) to \(D_a^{j}\) as illustrated in Fig.~\ref{fig:patch flow}b, where \(*_a^j\) denotes the j-th \textit{answer} position (or local task vector) of the sequence. 

\begin{table}[htb]
\caption{Tasks used in our study. Full input--output examples are provided in Appendix Table~\ref{tab:task_list_full}.}
\label{tab:task_list}
\centering
\small
\setlength{\tabcolsep}{6pt}
\begin{tabularx}{\linewidth}{lX}
\toprule
\textbf{Task Name} & \textbf{Task Rule} \\
\midrule
\multicolumn{2}{l}{\textit{Categorization tasks} ($y \in \{0,1\}$)}\\
Simple String  & $y=1$ iff $\mathrm{len}(s)>5$. \\
Complex String & $y=1$ iff $\mathrm{len}(s)>5$ (random chars). \\
Digit          & $y=1$ iff digit $\ge 5$. \\
2-D Data       & $y=1$ iff $y \ge x$ for input $(x,y)$. \\
\midrule
\multicolumn{2}{l}{\textit{Knowledge tasks}}\\
Antonyms   & Adjective $\rightarrow$ Antonym. \\
Capital    & Country $\rightarrow$ Capital. \\
Language   & Location $\rightarrow$ Language. \\
Profession & Person $\rightarrow$ Profession. \\
Religion   & Person $\rightarrow$ Religion. \\
\bottomrule
\end{tabularx}
\end{table}

\subsection{Accuracy Per Layer}
For comparison, we also show the results from patching the global task vector. Given an original sequence \(S\) and a dummy sequence \(D\), we patch \(\hat{S}_i^{j}\) to \(D_i^{0}\) as illustrated in Fig.~\ref{fig:patch flow}a, where \(*_i^j\) denotes the j-th \textit{is} position of the sequence and \(\hat{S}_i^{j}\) is the \textbf{averaged} task vector across samples described in the Preliminaries. This dummy sequence is zero-shot, i.e., it contains no demonstrations. When calculating the accuracy of each layer, we apply the final layer normalization and the language model head to the hidden states in these layers as in \cite{hendel2023incontext}. Patching with the local task vectors and the global task vector has different effects in the two tasks.

\begin{figure}[htb]
    \centering
    \includegraphics[width=0.7\linewidth]{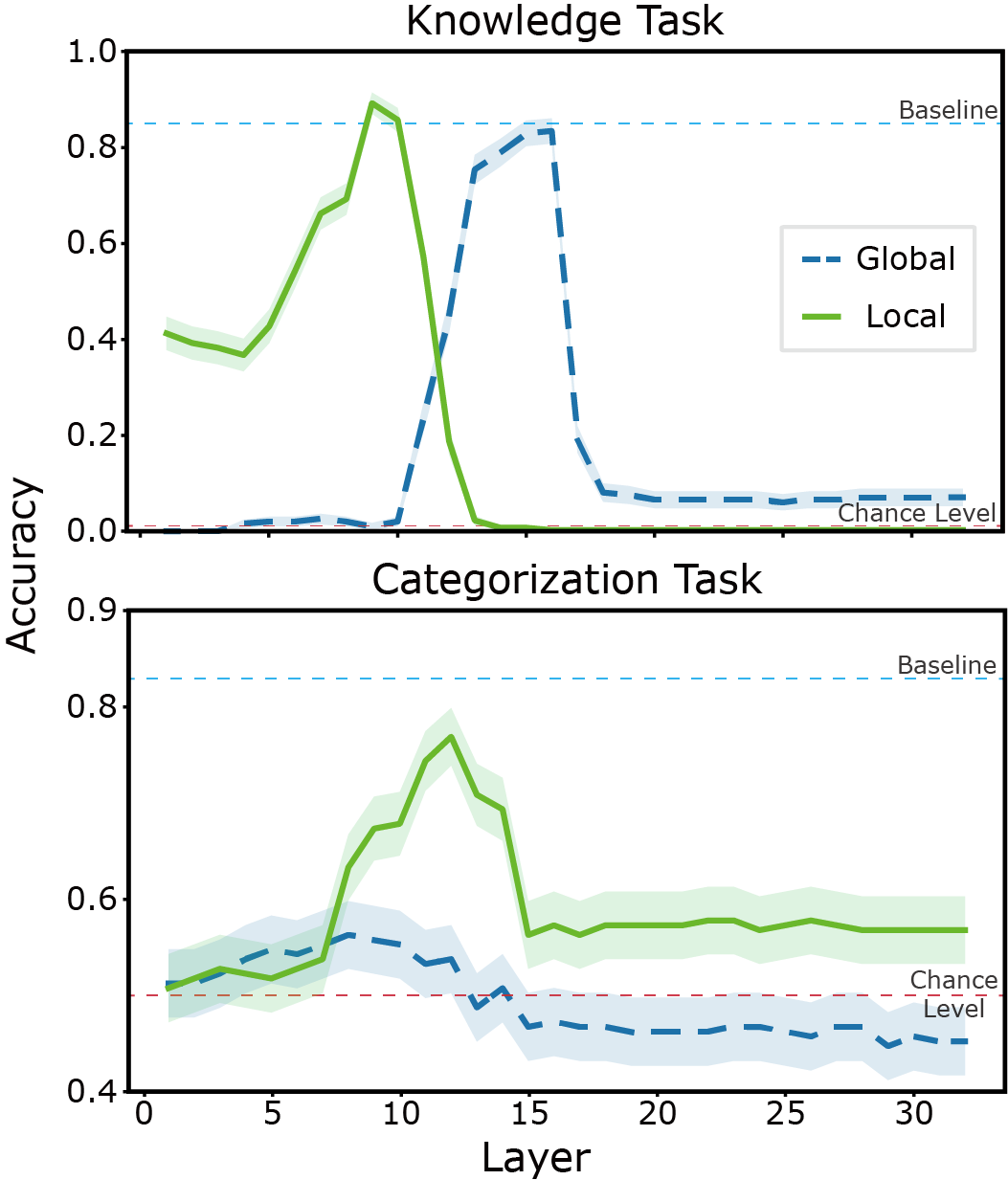}
    \caption{Patching with the global and local task vectors. The accuracy of the network patched with the \textit{global task vector} and the \textit{local task vectors} in different layers in the knowledge task (top) and the categorization task (bottom) is plotted against the layer number. Blue dashed lines indicate the baseline performance, which is the network receiving normal demonstrations. Red dashed lines indicate the chance level.}
    \label{fig:patchlayer}
\end{figure}

\begin{figure}[hb]
    \centering
    \includegraphics[width=0.7\linewidth]{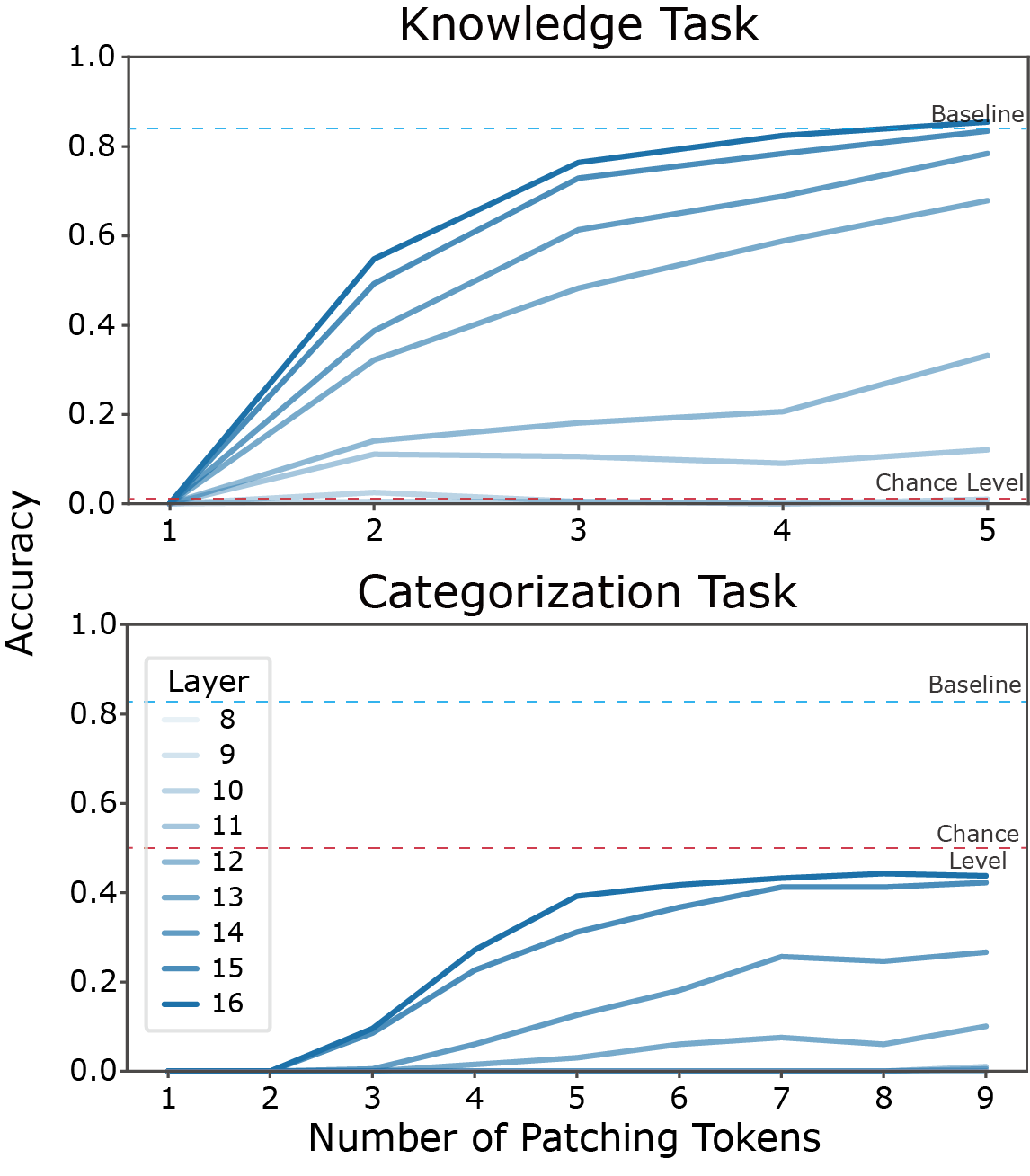}
    \caption{Patching the global task vectors in 8-16 layers. Top: knowledge task; Bottom: categorization task.}
    \label{fig:patch task vector acc}
\end{figure}

Patching with the global task vector restores the ICL performance with dummy inputs in the knowledge task. The layers where patching works the best coincide with the layers where we see the sharp increases in accuracy (Fig.~\ref{fig:acc_by_layers}).  Global task vectors created with more demonstrations lead to higher accuracy (Fig.~\ref{fig:patch task vector acc}). With global task vectors based on three or more demonstrations, the zero-shot performance of the patched LLM is on par with that receiving the real demonstrations. However, the global task vector does not restore the performance with dummy inputs in the categorization task. The patched LLM performs near the chance level, which is 50\% (Fig.~\ref{fig:patchlayer}). In these failed global-vector patching cases, the model often assigns highest probability to non-label tokens (i.e., outputs arbitrary tokens instead of \{\texttt{0}, \texttt{1}\}), consistent with an insufficient task representation.

\begin{figure}[ht]
    \centering
    \includegraphics[width=0.7\linewidth]{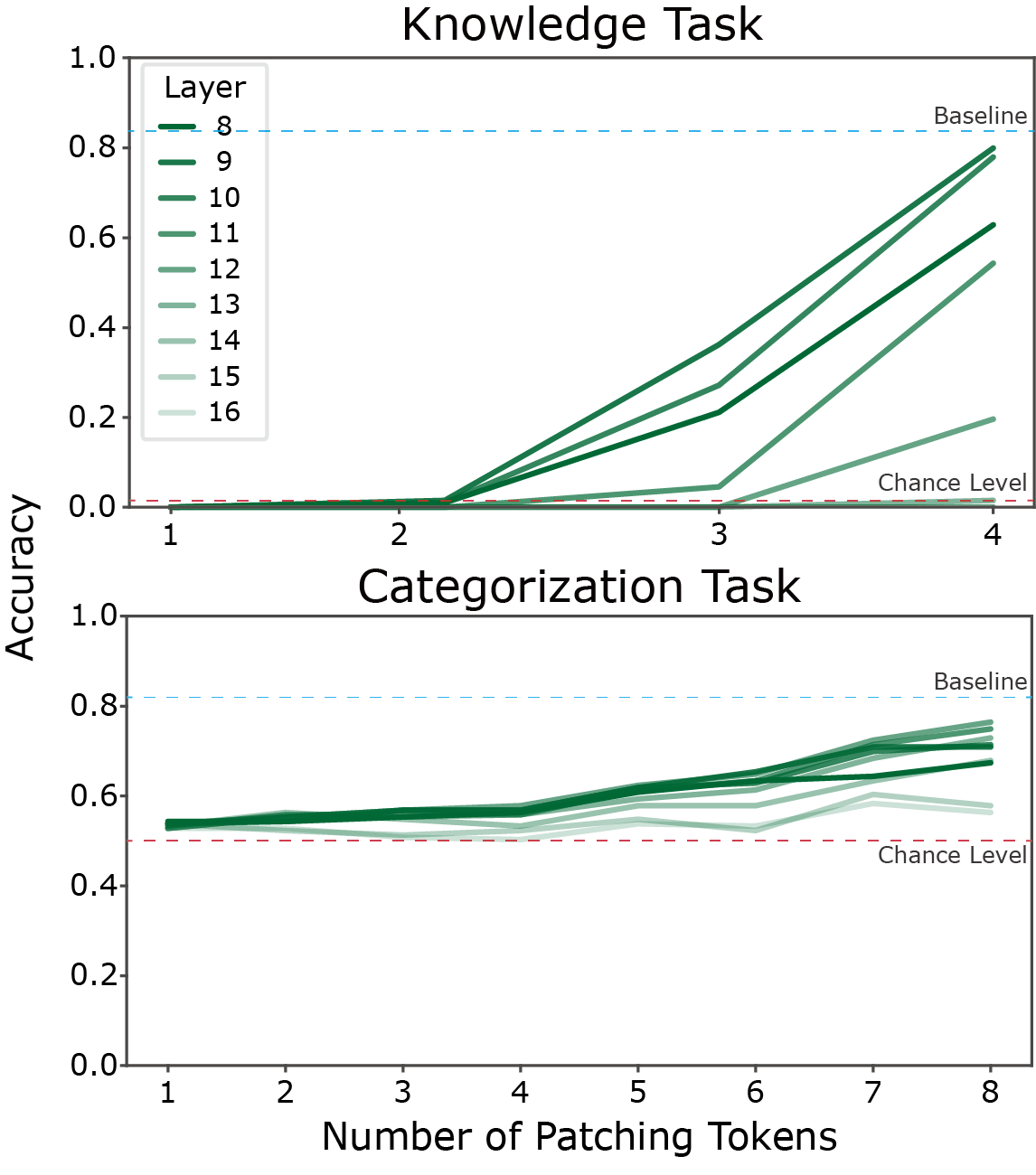}
    \caption{Patching the local task vectors in 8-16 layers. Top: knowledge task; Bottom: categorization task.}
    \label{fig:patch rule vector acc}
\end{figure}

In contrast, patching with the local task vectors in the middle layers rescues the performance of the model in the categorization task (Fig.~\ref{fig:patchlayer}, bottom). The more demonstrations' local task vectors are patched, the better the performance is (Fig.~\ref{fig:patch rule vector acc}). Interestingly, patching with the local task vectors also improves the LLM's performance in the knowledge task. 

In addition, the global and local task vectors work differently at different layers (Fig. ~\ref{fig:patchlayer}). We examine the LLM performance when we apply the patch at different layers and find that patching the local task vectors works best in earlier layers (6-10), and patching the global task vector works better in relatively later layers (12-15). This result supports the previous finding \cite{wang2023label} that the information contained within the local task vectors converges toward the global task vector in the later layers during the knowledge task. However, such translation of task information representation is not found in the categorization task (Fig.~\ref{fig:patchlayer} Bottom).

We carry out further patching experiments on various knowledge-based tasks and categorization tasks and on different LLMs. Main results are summarized in Table \ref{tab:main_task_acc}. We additionally validate the findings on larger and different model families; see Table~\ref{tab:main_task_acc_cont} and~\ref{tab:all_task_acc} in Appendix~\ref{sec:append3}. For knowledge-based tasks, we selected different types of factual tasks. The categorization tasks require multiple demonstrations to establish the correct association rules. We find that in some tasks, patching the global task vector to the dummy position leads to poor performance, especially in the categorization tasks. In cases where patching with the global task vector does not work well, patching with the local task vectors usually makes up for the performance.

Besides, patching with the global task vector is often more effective when we calculate the global task vector by averaging across the dataset, which makes it not just global for a particular question, but also global in the sense of the whole dataset. This method may extract the most accurate task information. However, in many categorization tasks, local task vectors often outperform the global task vector, even with averaging. Moreover, averaging across datasets is not feasible in normal inference. Therefore, our findings on the local task vector may unveil a more general mechanism underlying In-Context Learning.

Across different architectures, we observe a consistent trend where local task vectors outperform global ones in categorization tasks. Notably, Vicuna-7B shows strong performance with local vector patching in the Simple String and Digit tasks, highlighting the robustness of this mechanism across model families.

\section{Limitations}
\label{sec:limitation}
Neither global nor local task vectors work consistently across all tasks and in all types of LLMs tested. We are yet to determine the mechanism causing this inconsistency. Although patching local task vectors works well in most tasks, it fails to restore the few-shot performance in a small number of models and tasks, and the reason for this has yet to be explored.

\section{Conclusion}
Our results reveal a more general mechanism involving distributed local task vectors that encapsulate the task information contained in the demonstrations during ICL. The local task vectors encode the task information in an abstract manner. Patching the local task vectors to models receiving dummy inputs yields performance levels comparable to few-shot learning. Notably, local task vectors are present in certain tasks where global task vectors are absent, particularly in categorization tasks. This suggests a more nuanced, distributed approach to processing information and extracting task rules in ICL.

\begin{table*}[t]
    \caption{Best accuracy after patching global vs.\ local task vectors on \textsc{LLaMA}-7B across tasks. Baseline is standard ICL with real demonstrations; patched models receive dummy demonstrations. Averaging settings follow Sec.~\ref{sec:eval_details}. Results for other models are reported in Appendix Table~\ref{tab:main_task_acc_cont} and Table~\ref{tab:all_task_acc}. Results are averaged over 5 random trials; variances are small and omitted for brevity.}
    \label{tab:main_task_acc} 
    \centering
    \setlength{\tabcolsep}{3pt} 
    \begin{tabular}{@{}ccccccccc@{}}
    \toprule
    \multirow{2}{*}{\textbf{Model}} & \multirow{2}{*}{\textbf{Task Name}} & \multirow{2}{*}{\textbf{1-shot}} & \multirow{2}{*}{\textbf{Baseline}} & \multicolumn{2}{c}{\textbf{Global Task Vector}} & \multicolumn{3}{c}{\textbf{Local Task Vector}} \\
     & & & & No Avg & Trial Avg  & No Avg & Trial Avg & Pos Avg \\
    \midrule
    \multirow{14}{*}{LLaMA-7B} & \multirow{2}{*}{\shortstack{Simple \\ String}} & \multirow{2}{*}{0.37} & \multirow{2}{*}{0.79} & \multirow{2}{*}{0.49} & \multirow{2}{*}{0.63} & \multirow{2}{*}{\textbf{0.66}} & \multirow{2}{*}{0.58} & \multirow{2}{*}{\textbf{0.66}} \\
     &  &  &  &  &  &  &  &  \\
     & \multirow{2}{*}{\shortstack{Complex \\ String}} &  \multirow{2}{*}{0.57} & \multirow{2}{*}{0.80} & \multirow{2}{*}{0.55} & \multirow{2}{*}{0.56} & \multirow{2}{*}{\textbf{0.65}} & \multirow{2}{*}{0.61} & \multirow{2}{*}{0.39} \\
     &  &  &  &  &  &  &  &  \\
     & Digit & 0.43 & 0.69 & 0.63 & 0.57 & 0.65 & \textbf{0.68} & 0.60 \\
     & \multirow{2}{*}{\shortstack{2-D \\ 8 Demo}} & \multirow{2}{*}{0.00} & \multirow{2}{*}{0.59} & \multirow{2}{*}{0.53} & \multirow{2}{*}{0.49} & \multirow{2}{*}{\textbf{0.59}} & \multirow{2}{*}{0.52} & \multirow{2}{*}{0.38} \\
     &  &  &  &  &  &  &  &  \\
     & \multirow{2}{*}{\shortstack{2-D \\ 16 Demo}} & \multirow{2}{*}{-} & \multirow{2}{*}{0.67} & \multirow{2}{*}{0.53} & \multirow{2}{*}{0.49} & \multirow{2}{*}{\textbf{0.55}} & \multirow{2}{*}{0.52} & \multirow{2}{*}{0.43} \\
     &  &  &  &  &  &  &  &  \\
    \cline{2-9}
     & Antonyms & 0.56 & 0.86 & 0.75 & 0.79 & \textbf{0.81} & 0.80 & 0.00 \\
     & Capital & 0.68 & 0.86 & 0.83 & \textbf{0.92} & 0.80 & 0.83 & 0.00 \\
     & Language & 0.67 & 0.82 & 0.64 & 0.79 & 0.77 & \textbf{0.85} & 0.00 \\
     & Profession & 0.35 & 0.54 & 0.32 & \textbf{0.53} & 0.48 & 0.51 & 0.00 \\			
     & Religion & 0.60 & 0.87 & 0.49 & 0.71 & 0.80 & \textbf{0.92} & 0.00 \\
    \bottomrule
    \end{tabular}
\end{table*}

\section{Related Work}
\subsection{Neuroscience}
Our work is inspired by studies from the field of neuroscience \cite{barrett2019analyzing, richards2019deep, yamins2016using, yousefi2023decoding} that explore the computational mechanisms underlying cognitive functions of the brain. The current study adopts a categorization task similar to what has been used widely in testing animals and humans in neuroscience. These experiments typically provide the demonstrations sequentially. Subjects learn the task through trial-and-error. This process is typically modeled as reinforcement learning \cite{niv2009reinforcement}. However, it has been shown that a system that stores behavior history could model the learning equally well without using explicit reinforcement learning \cite{zhang2018neural}. This is conceptually similar to ICL in LLMs. The demixed-PCA (dPCA) \cite{kobak2016demixed} used in our study is also initially developed to study how information is encoded in a high-dimensional space represented by population neuronal activities.

\subsection{In-Context Learning}
\citet{brown2020language} discovered that LLMs possess few-shot learning abilities, enabling them to perform reasoning across different tasks from just a few demonstrations. Meta-learning capabilities may be behind the models' capability to efficiently adapt to new information \cite{dai2023why}. It has been further proposed that transformers utilize gradient descent to perform linear regression tasks \cite{ahn2024transformers, akyurek2023what, von2023transformers}. Another approach to understanding ICL in LLMs is to view pre-trained models as implicit Bayesian models, and the demonstrations provided in the prompts allow the model to compute the posteriors \cite{ahuja2023context, wang2023large, xie2022explanation}.
Mechanistic analyses further link few-shot ICL to concrete circuits such as induction heads, whose ablation can substantially reduce the benefit from demonstrations~\cite{crosbie2025induction}.
Our work complements these circuit-level analyses by providing a spatial perspective on how information aggregates across different tokens.

\subsection{Task Vector}
It is proposed that LLMs might be either compressing abstract rule information from demonstrations \cite{hendel2023incontext} or aggregating demonstration information layer-by-layer \cite{wang2023label}. An example is to tell the capital of a country. We term these tasks knowledge tasks. Reference \cite{hendel2023incontext} argued that ICL is achieved through a single task vector in the middle layers extracted from the demonstrations. 
Recent evidence also suggests that complex ICL may rely on multiple task-relevant components rather than a single averaged vector~\cite{saglam2025learning, tikhonov2025one}, which our local-vs-global analysis mechanistically contextualizes.

\bibliography{References.bib}

\appendix

\section{Appendix: Supplemental Material}

\subsection{Analysis Details}
\label{sec:details}

\paragraph{Models and Datasets}
We employ the LLaMA from HuggingFace \cite{touvron2023LLaMA}  as our primary model, and the model weights are also sourced from HuggingFace. No further processing on the weights is performed. 

We also use the Pythia \cite{biderman2023Pythia}, Vicuna \cite{chiang2023vicuna}, LLaMA 3~\cite{dubey2024llama} and Qwen 2.5~\cite{qwen2024qwen2} models. Results on LLaMA-7B are reported in Table~\ref{tab:main_task_acc}; results on other models are reported in Tables~\ref{tab:main_task_acc_cont} and~\ref{tab:all_task_acc}.

The task formats listed in Table 1 follow existing task definitions used in prior works such as ~\citet{hendel2023incontext} and ~\citet{wang2023label}. These references provide detailed descriptions for both factual knowledge and synthetic categorization tasks.

\paragraph{Tokenization}
Our experiments require us to retrieve tokens at specific positions. However, the tokenizer used with LLaMA generates variable numbers of tokens even for strings of the same length. Therefore, we insert a special character, e.g., “\(\sim\)”, into the original input text to force the tokenizer to generate the same number of tokens for strings of the same length. We remove the token of these special characters after the tokenization step.

We verified that the inclusion of the special character ``$\sim$'' for length normalization does not introduce inductive biases or alter model performance. Inference results with and without ``$\sim$'' are identical across all benchmarked tasks, confirming the robustness of this strategy.

\paragraph{Evaluation Details}
\label{sec:eval_details}
We use 5000 samples for calculating network performance. PCA and dPCA analyses are done with 1000 samples. More samples produce similar results. All experiments are done with a single V100 GPU. Accuracy is evaluated with the token with the largest logit without temperature or any other probabilistic sampling methods.For binary categorization tasks, a prediction is counted as correct only if the top-1 token matches the label token (\texttt{0} or \texttt{1}); outputs outside \{\texttt{0}, \texttt{1}\} are counted as incorrect.

To ensure the reliability of our results, all experiments were conducted across 5 random seeds for demonstration selection and ordering. We observed minimal variance (standard deviation $< 0.01$ for most tasks), and the performance gaps between local and global patching are statistically significant ($p < 0.05$ under a paired t-test).

We denote the three types of averaging strategies in Table~\ref{tab:main_task_acc} as follows: \textbf{No Avg} indicates that the result is from a single trial without averaging across runs. \textbf{Trial Avg} refers to averaging results of the same position across multiple trials. \textbf{Pos Avg} represents averaging across different positions within a single trial. These settings allow us to assess the consistency and locality of task vector effectiveness.

\subsection{Ablation at Answer Position}\label{sec:append2}
In Figure~\ref{fig:lesioning}, we perform an ablation experiment by setting the tokens at the answer positions of the demonstrations to zero. For the knowledge task, the ablation of the answer tokens across all demonstrations is necessary for a significant effect. In contrast, for the categorization task, the performance decreases gradually as the number of ablated answer tokens increases, suggesting a distributed mechanism. This gradual decline indicates that task information is spread across multiple tokens, highlighting the robustness of the model in categorization tasks even when partial information is removed. 

Additionally, we performed preliminary tests by shuffling the order of demonstrations within the prompt. We observed that demonstration ordering had minimal impact on the qualitative trends reported in Section~\ref{sec:patching_vector} (Fig.~\ref{fig:patchlayer}, \ref{fig:patch task vector acc} and \ref{fig:patch rule vector acc}), further supporting the distributed nature of the identified local task vectors.

\subsection{Table of Experiment}\label{sec:append3}
We conduct additional experiments across several models and tasks and report the performance both with and without averaging across samples. For the local task vectors, we perform extra experiments by averaging across the demonstration, which leads to a poorer performance. The result suggests that the information contained in different local task vectors cannot be simply averaged. It further supports the distributed nature of these vectors.

\subsection{Licenses for existing assets}
We follow the license of LLaMA (\href{https://github.com/Meta-LLaMA/LLaMA/blob/main/LICENSE}{LLaMA License}), Pythia (\href{https://github.com/EleutherAI/Pythia/blob/main/LICENSE}{Pythia License}) and Vicuna (\href{https://github.com/vproc/vicuna/blob/main/LICENSE.txt}{Vicuna License}) and Apache License 2.0 for HuggingFace.

\begin{table*}[t]
\caption{Full task formats with input--output examples (supplement to Table~\ref{tab:task_list}).}
\label{tab:task_list_full}
\centering
\footnotesize
\setlength{\tabcolsep}{4pt}
\begin{tabularx}{\textwidth}{lX X}
\toprule
\textbf{Task Name} & \textbf{Task Rule} & \textbf{Example} \\
\midrule
\multicolumn{3}{l}{\textit{Categorization tasks} ($y \in \{0,1\}$)}\\
Simple String  & 1 if len$>5$ else 0 & abaabb$\to$1, aab$\to$0 \\
Complex String & 1 if len$>5$ else 0 & aFeXGb$\to$1, axvb$\to$0 \\
Digit          & 1 if digit$\ge$5 else 0 & 4$\to$0, 7$\to$1 \\
2-D Data       & 1 if $y\ge x$ else 0 & (0,1)$\to$1, (7,4)$\to$0 \\
\midrule
\multicolumn{3}{l}{\textit{Knowledge tasks}}\\
Antonyms   & Given an English adjective, output an antonym & Adjective$\to$Antonym \\
Capital    & Given a country name, output its capital city  & Country$\to$Capital \\
Language   & Given a location name, output its native language & Location$\to$Language \\
Profession & Given a person name, output their profession & Person$\to$Profession \\
Religion   & Given a person name, output the associated religion & Person$\to$Religion \\
\bottomrule
\end{tabularx}
\end{table*}

\begin{table*}[t]
\caption{(continued) Results on other 7B-scale models.}
\label{tab:main_task_acc_cont} 
\centering
\setlength{\tabcolsep}{3pt} 
\begin{tabular}{@{}ccccccccc@{}}
\toprule
\multirow{2}{*}{\textbf{Model}} & \multirow{2}{*}{\textbf{Task Name}} & \multirow{2}{*}{\textbf{1-shot}} & \multirow{2}{*}{\textbf{Baseline}} & \multicolumn{2}{c}{\textbf{Global Task Vector}} & \multicolumn{3}{c}{\textbf{Local Task Vector}} \\
 & & & & No Avg & Trial Avg  & No Avg & Trial Avg & Pos Avg \\
\midrule
\multirow{14}{*}{Pythia-6.9B} & \multirow{2}{*}{\shortstack{Simple \\ String}} & \multirow{2}{*}{0.37} & \multirow{2}{*}{0.80} & \multirow{2}{*}{0.47} & \multirow{2}{*}{0.59} & \multirow{2}{*}{0.58} & \multirow{2}{*}{\textbf{0.61}} & \multirow{2}{*}{0.00} \\
 &  &  &  &  &  &  &  &  \\
 & \multirow{2}{*}{\shortstack{Complex \\ String}} &  \multirow{2}{*}{0.52} & \multirow{2}{*}{0.69} & \multirow{2}{*}{0.53} & \multirow{2}{*}{\textbf{0.55}} & \multirow{2}{*}{0.49} & \multirow{2}{*}{0.51} & \multirow{2}{*}{0.00} \\
 &  &  &  &  &  &  &  &  \\
 & Digit & 0.55& 0.66& 0.58& 0.55& \textbf{0.60}& 0.52& 0.00 \\
 & \multirow{2}{*}{\shortstack{2-D \\ 8 Demo}} & \multirow{2}{*}{0.03} & \multirow{2}{*}{0.63} & \multirow{2}{*}{0.55} & \multirow{2}{*}{0.50} & \multirow{2}{*}{\textbf{0.59}} & \multirow{2}{*}{0.53} & \multirow{2}{*}{0.00} \\
 &  &  &  &  &  &  &  &  \\
 & \multirow{2}{*}{\shortstack{2-D \\ 16 Demo}} & \multirow{2}{*}{-} & \multirow{2}{*}{0.71} & \multirow{2}{*}{0.55} & \multirow{2}{*}{0.50} & \multirow{2}{*}{\textbf{0.59}} & \multirow{2}{*}{\textbf{0.59}} & \multirow{2}{*}{0.00} \\
 &  &  &  &  &  &  &  &  \\
\cline{2-9}
 & Antonyms & 0.26& 0.84& 0.75& 0.80& \textbf{0.81}& 0.80& 0.01 \\
 & Capital  & 0.68& 0.81& 0.79& \textbf{0.85}& 0.73& 0.83& 0.01 \\
 & Language & 0.45& 0.66& 0.62& \textbf{0.69}& 0.51& 0.68& 0.00 \\
 & Profession & 0.2& 0.23& 0.27& \textbf{0.47}& 0.33& 0.39& 0.00 \\
 & Religion & 0.60& 0.84& 0.49& \textbf{0.87}& 0.83& 0.85& 0.00\\
\midrule
\multirow{14}{*}{Vicuna-7B} & \multirow{2}{*}{\shortstack{Simple \\ String}} & \multirow{2}{*}{0.37} & \multirow{2}{*}{0.71} & \multirow{2}{*}{0.48}  & \multirow{2}{*}{0.63} & \multirow{2}{*}{0.67} & \multirow{2}{*}{0.58} & \multirow{2}{*}{\textbf{0.70}} \\
 &  &  &  &  &  &  &  &  \\
 & \multirow{2}{*}{\shortstack{Complex \\ String}} &  \multirow{2}{*}{0.57} & \multirow{2}{*}{0.71} & \multirow{2}{*}{0.59} & \multirow{2}{*}{0.56} & \multirow{2}{*}{\textbf{0.62}} & \multirow{2}{*}{0.58} & \multirow{2}{*}{0.09} \\
 &  &  &  &  &  &  &  &  \\
 & Digit & 0.55& 0.71& 0.64& \textbf{0.78}& 0.64& 0.66& 0.17 \\
 & \multirow{2}{*}{\shortstack{2-D \\ 8 Demo}} & \multirow{2}{*}{0.00} & \multirow{2}{*}{0.57} & \multirow{2}{*}{0.59} & \multirow{2}{*}{\textbf{0.80}} & \multirow{2}{*}{0.57} & \multirow{2}{*}{0.52} & \multirow{2}{*}{0.03} \\
 &  &  &  &  &  &  &  &  \\
 & \multirow{2}{*}{\shortstack{2-D \\ 16 Demo}} & \multirow{2}{*}{-} & \multirow{2}{*}{0.58} & \multirow{2}{*}{0.59} & \multirow{2}{*}{\textbf{0.80}} & \multirow{2}{*}{0.55} & \multirow{2}{*}{0.55} & \multirow{2}{*}{0.03} \\
 &  &  &  &  &  &  &  &  \\
\cline{2-9}
 & Antonyms & 0.47& 0.83& 0.56& 0.77 & \textbf{0.84}& 0.83& 0.00 \\
 & Capital  & 0.63& 0.87& 0.76& \textbf{0.92} & 0.77& 0.85& 0.00 \\
 & Language & 0.5& 0.78& 0.49& 0.70& 0.75& \textbf{0.83}& 0.00 \\
 & Profession & 0.32& 0.44& 0.28& \textbf{0.53}& 0.45& 0.52& 0.00 \\
 & Religion & 0.72& 0.88& 0.50& \textbf{0.94}& 0.86& 0.91& 0.00 \\
\bottomrule
\end{tabular}
\end{table*}

\begin{table*}[t]
\ContinuedFloat
\caption{(continued) Results on other 7B-scale models.}
\label{tab:main_task_acc_cont_llama3_qwen} 
\centering
\setlength{\tabcolsep}{3pt} 
\begin{tabular}{@{}ccccccccc@{}}
\toprule
\multirow{2}{*}{\textbf{Model}} & \multirow{2}{*}{\textbf{Task Name}} & \multirow{2}{*}{\textbf{1-shot}} & \multirow{2}{*}{\textbf{Baseline}} & \multicolumn{2}{c}{\textbf{Global Task Vector}} & \multicolumn{3}{c}{\textbf{Local Task Vector}} \\
 & & & & No Avg & Trial Avg  & No Avg & Trial Avg & Pos Avg \\
\midrule

\multirow{14}{*}{LLaMA3-8B}
 & \multirow{2}{*}{\shortstack{Simple \\ String}} & \multirow{2}{*}{0.42} & \multirow{2}{*}{0.83} & \multirow{2}{*}{0.63} & \multirow{2}{*}{0.65} & \multirow{2}{*}{\textbf{0.72}} & \multirow{2}{*}{0.66} & \multirow{2}{*}{0.60} \\
 &  &  &  &  &  &  &  &  \\
 & \multirow{2}{*}{\shortstack{Complex \\ String}} & \multirow{2}{*}{0.60} & \multirow{2}{*}{0.84} & \multirow{2}{*}{0.60} & \multirow{2}{*}{0.62} & \multirow{2}{*}{\textbf{0.70}} & \multirow{2}{*}{0.64} & \multirow{2}{*}{0.55} \\
 &  &  &  &  &  &  &  &  \\
 & Digit & 0.50 & 0.75 & 0.60 & \textbf{0.68} & \textbf{0.68} & 0.62 & 0.58 \\
 & \multirow{2}{*}{\shortstack{2-D \\ 8 Demo}} & \multirow{2}{*}{0.04} & \multirow{2}{*}{0.66} & \multirow{2}{*}{0.52} & \multirow{2}{*}{0.54} & \multirow{2}{*}{\textbf{0.61}} & \multirow{2}{*}{0.56} & \multirow{2}{*}{0.45} \\
 &  &  &  &  &  &  &  &  \\
 & \multirow{2}{*}{\shortstack{2-D \\ 16 Demo}} & \multirow{2}{*}{-} & \multirow{2}{*}{0.73} & \multirow{2}{*}{0.54} & \multirow{2}{*}{0.56} & \multirow{2}{*}{\textbf{0.66}} & \multirow{2}{*}{0.60} & \multirow{2}{*}{0.48} \\
 &  &  &  &  &  &  &  &  \\
\cline{2-9}
 & Antonyms  & 0.60 & 0.90 & 0.86 & 0.85 & \textbf{0.88} & 0.87 & 0.01 \\
 & Capital   & 0.72 & 0.92 & 0.90 & \textbf{0.91} & 0.88 & 0.89 & 0.00 \\
 & Language  & 0.70 & 0.88 & 0.83 & 0.84 & \textbf{0.86} & 0.85 & 0.00 \\
 & Profession& 0.40 & 0.62 & 0.56 & \textbf{0.59} & 0.58 & 0.58 & 0.00 \\
 & Religion  & 0.65 & 0.91 & 0.87 & 0.89 & 0.88 & \textbf{0.90} & 0.00 \\

\midrule

\multirow{14}{*}{Qwen2.5-7B}
 & \multirow{2}{*}{\shortstack{Simple \\ String}} & \multirow{2}{*}{0.46} & \multirow{2}{*}{0.85} & \multirow{2}{*}{0.66} & \multirow{2}{*}{0.67} & \multirow{2}{*}{\textbf{0.74}} & \multirow{2}{*}{0.68} & \multirow{2}{*}{0.62} \\
 &  &  &  &  &  &  &  &  \\
 & \multirow{2}{*}{\shortstack{Complex \\ String}} & \multirow{2}{*}{0.62} & \multirow{2}{*}{0.86} & \multirow{2}{*}{0.62} & \multirow{2}{*}{0.63} & \multirow{2}{*}{\textbf{0.72}} & \multirow{2}{*}{0.66} & \multirow{2}{*}{0.56} \\
 &  &  &  &  &  &  &  &  \\
 & Digit & 0.52 & 0.78 & 0.61 & 0.63 & \textbf{0.71} & 0.64 & 0.60 \\
 & \multirow{2}{*}{\shortstack{2-D \\ 8 Demo}} & \multirow{2}{*}{0.06} & \multirow{2}{*}{0.68} & \multirow{2}{*}{0.54} & \multirow{2}{*}{\textbf{0.63}} & \multirow{2}{*}{0.62} & \multirow{2}{*}{0.57} & \multirow{2}{*}{0.46} \\
 &  &  &  &  &  &  &  &  \\
 & \multirow{2}{*}{\shortstack{2-D \\ 16 Demo}} & \multirow{2}{*}{-} & \multirow{2}{*}{0.75} & \multirow{2}{*}{0.56} & \multirow{2}{*}{0.57} & \multirow{2}{*}{\textbf{0.67}} & \multirow{2}{*}{0.61} & \multirow{2}{*}{0.50} \\
 &  &  &  &  &  &  &  &  \\
\cline{2-9}
 & Antonyms  & 0.62 & 0.91 & 0.88 & \textbf{0.90} & 0.87 & 0.89 & 0.01 \\
 & Capital   & 0.75 & 0.93 & 0.91 & \textbf{0.92} & 0.89 & 0.90 & 0.00 \\
 & Language  & 0.72 & 0.90 & 0.86 & 0.88 & 0.87 & \textbf{0.89} & 0.00 \\
 & Profession& 0.42 & 0.65 & 0.60 & \textbf{0.62} & 0.59 & 0.61 & 0.00 \\
 & Religion  & 0.68 & 0.92 & 0.88 & 0.90 & 0.89 & \textbf{0.91} & 0.00 \\

\bottomrule
\end{tabular}
\end{table*}

\begin{table*}[ht]
    \caption{Additional results on other model sizes (same setup as Table~\ref{tab:main_task_acc}).}
    \label{tab:all_task_acc} 
    \centering
    \footnotesize
    \begin{tabular}{ccccccccc}
        \toprule
        \multirow{2}{*}{\textbf{Model}} & \multirow{2}{*}{\textbf{Task Name}} & \multirow{2}{*}{\textbf{1-shot}} & \multirow{2}{*}{\textbf{Baseline}} & \multicolumn{2}{c}{\textbf{Global Task Vector}} & \multicolumn{3}{c}{\textbf{Local Task Vector}} \\
         & & & & No Avg & Trial Avg  & No Avg & Trial Avg & Pos Avg \\
							
         \midrule
         \multirow{14}{*}{LLaMA-13B} & \multirow{2}{*}{\shortstack{Simple \\ String}} & \multirow{2}{*}{0.00} & \multirow{2}{*}{0.72} & \multirow{2}{*}{0.45} & \multirow{2}{*}{0.63} & \multirow{2}{*}{\textbf{0.67}} & \multirow{2}{*}{0.64} & \multirow{2}{*}{0.28} \\				
         &  &  &  &  &  &  &  &  \\
         & \multirow{2}{*}{\shortstack{Complex \\ String}} &  \multirow{2}{*}{0.00} & \multirow{2}{*}{0.77} & \multirow{2}{*}{0.52} & \multirow{2}{*}{0.58} & \multirow{2}{*}{\textbf{0.66}} & \multirow{2}{*}{0.62} & \multirow{2}{*}{0.27} \\	
         &  &  &  &  &  &  &  &  \\
         & Digit & 0.61 & 0.54 & 0.51 & \textbf{0.62} & 0.53 & 0.53 & 0.53 \\
         & \multirow{2}{*}{\shortstack{2-D \\ 8 Demo}} & \multirow{2}{*}{0.00} & \multirow{2}{*}{0.57} & \multirow{2}{*}{\textbf{0.60}} & \multirow{2}{*}{0.49} & \multirow{2}{*}{0.53} & \multirow{2}{*}{0.52} & \multirow{2}{*}{0.31} \\
         &  &  &  &  &  &  &  &  \\
         & \multirow{2}{*}{\shortstack{2-D \\ 16 Demo}} & \multirow{2}{*}{-} & \multirow{2}{*}{0.56} & \multirow{2}{*}{\textbf{0.61}} & \multirow{2}{*}{0.49} & \multirow{2}{*}{0.53} & \multirow{2}{*}{0.48} & \multirow{2}{*}{0.26} \\
         &  &  &  &  &  &  &  &  \\
         \cline{2-9}
         & Antonyms &0.61	& 0.77 & 0.65 & 0.78 & \textbf{0.85} & \textbf{0.85} &	0.09 \\
         & Capital & 0.74&0.84	&	0.82	&\textbf{0.88}	&0.80&	0.83&0.05\\
         & Language &0.56&	0.77&		0.56&	\textbf{0.85}&	0.75&	0.79&	0.01 \\
         & Profession &0.27&	0.44	&	0.35&	\textbf{0.49}&	0.40&	0.48&	0.03 \\
         & Religion &0.57	&0.75	&	0.53&	0.80&	0.82&	\textbf{0.84}&	0.00 \\
         
         \midrule
         
         \multirow{14}{*}{Pythia-2.8B} & \multirow{2}{*}{\shortstack{Simple \\ String}} & \multirow{2}{*}{0.37} & \multirow{2}{*}{0.81} & \multirow{2}{*}{0.48} & \multirow{2}{*}{0.70} & \multirow{2}{*}{\textbf{0.72}} & \multirow{2}{*}{0.64} & \multirow{2}{*}{0.00} \\		
         &  &  &  &  &  &  &  &  \\
         & \multirow{2}{*}{\shortstack{Complex \\ String}} &  \multirow{2}{*}{0.57} & \multirow{2}{*}{0.73} & \multirow{2}{*}{0.55} & \multirow{2}{*}{0.70} & \multirow{2}{*}{0.65} & \multirow{2}{*}{\textbf{0.72}} & \multirow{2}{*}{0.00} \\
         &  &  &  &  &  &  &  &  \\
         & Digit & 0.55	&	0.72&	0.52&	0.55&	\textbf{0.72}&	0.69&	0.00\\
         & \multirow{2}{*}{\shortstack{2-D \\ 8 Demo}} & \multirow{2}{*}{0.10} & \multirow{2}{*}{0.67} & \multirow{2}{*}{0.55} & \multirow{2}{*}{0.50} & \multirow{2}{*}{\textbf{0.60}} & \multirow{2}{*}{0.56} & \multirow{2}{*}{0.00} \\
         &  &  &  &  &  &  &  &  \\
         & \multirow{2}{*}{\shortstack{2-D \\ 16 Demo}} & \multirow{2}{*}{-} & \multirow{2}{*}{0.73} & \multirow{2}{*}{0.55} & \multirow{2}{*}{0.50} & \multirow{2}{*}{\textbf{0.60}} & \multirow{2}{*}{\textbf{0.60}} & \multirow{2}{*}{0.00} \\
         &  &  &  &  &  &  &  &  \\
         \cline{2-9}
         & Antonyms & 0.05&	0.70&		0.09&	0.13&	0.63&	\textbf{0.73}&	0.00 \\
         & Capital & 0.58&	0.78&		0.78&	\textbf{0.82}&	0.68&	0.80	&0.01 \\
         & Language & 0.32&	0.70&		0.62&\textbf{0.76}&	0.64&	0.72&	0.01 \\
         & Profession & 0.20&0.25&		0.29&	\textbf{0.42}&	0.36&\textbf{0.42}&	0.00 \\
         & Religion &0.44&	0.84&		0.62&	0.75&	0.85&\textbf{0.89}&	0.00\\
         
         \midrule

          \multirow{14}{*}{Pythia-12B} & \multirow{2}{*}{\shortstack{Simple \\ String}} & \multirow{2}{*}{0.37} & \multirow{2}{*}{0.68} & \multirow{2}{*}{0.51} & \multirow{2}{*}{\textbf{0.72}} & \multirow{2}{*}{0.59} & \multirow{2}{*}{0.56} & \multirow{2}{*}{0.00} \\
         &  &  &  &  &  &  &  &  \\
         & \multirow{2}{*}{\shortstack{Complex \\ String}} &  \multirow{2}{*}{0.56} & \multirow{2}{*}{0.67} & \multirow{2}{*}{0.55} & \multirow{2}{*}{0.57} & \multirow{2}{*}{0.59} & \multirow{2}{*}{\textbf{0.62}} & \multirow{2}{*}{0.00} \\
         &  &  &  &  &  &  &  &  \\
         & Digit & 0.43&		0.67&	0.57&	0.55&	\textbf{0.62}&	0.61&	0.00\\
         & \multirow{2}{*}{\shortstack{2-D \\ 8 Demo}} & \multirow{2}{*}{0.02} & \multirow{2}{*}{0.64} & \multirow{2}{*}{0.54} & \multirow{2}{*}{\textbf{0.61}} & \multirow{2}{*}{0.53} & \multirow{2}{*}{0.57} & \multirow{2}{*}{0.00} \\
         &  &  &  &  &  &  &  &  \\
         & \multirow{2}{*}{\shortstack{2-D \\ 16 Demo}} & \multirow{2}{*}{-} & \multirow{2}{*}{0.64} & \multirow{2}{*}{0.53} & \multirow{2}{*}{\textbf{0.61}} & \multirow{2}{*}{0.53} & \multirow{2}{*}{0.53} & \multirow{2}{*}{0.00} \\
         &  &  &  &  &  &  &  &  \\
         \cline{2-9}
         & Antonyms & 0.12&	0.55&		0.13&	0.14&	\textbf{0.68}&	0.67&	0.00 \\
         & Capital & 0.12&	0.28&		0.29&	0.50&	0.39&	\textbf{0.68}&	0.00 \\
         & Language & 0.37&	0.69&		0.34&	0.56&	0.67&	\textbf{0.77}&	0.00\\
         & Profession & 0.07&	0.13&		0.18&	0.25&	0.37&	\textbf{0.40}&	0.00 \\
         & Religion & 0.26&	0.75&		0.46&	0.78&	0.85&	\textbf{0.92}&	0.00 \\

         \midrule
         
         \multirow{14}{*}{Vicuna-13B} & \multirow{2}{*}{\shortstack{Simple \\ String}} & \multirow{2}{*}{0.00} & \multirow{2}{*}{0.53} & \multirow{2}{*}{0.40} & \multirow{2}{*}{0.37} & \multirow{2}{*}{\textbf{0.53}} & \multirow{2}{*}{0.51} & \multirow{2}{*}{0.03} \\
         &  &  &  &  &  &  &  &  \\
         & \multirow{2}{*}{\shortstack{Complex \\ String}} &  \multirow{2}{*}{0.00} & \multirow{2}{*}{0.34} & \multirow{2}{*}{0.40} & \multirow{2}{*}{\textbf{0.56}} & \multirow{2}{*}{0.39} & \multirow{2}{*}{0.41} & \multirow{2}{*}{0.02} \\
         &  &  &  &  &  &  &  & \\
         & Digit & 0.13&		0.37&	0.48&	\textbf{0.55}&	0.45&	0.45&	0.05 \\
         & \multirow{2}{*}{\shortstack{2-D \\ 8 Demo}} & \multirow{2}{*}{0.00} & \multirow{2}{*}{0.36} & \multirow{2}{*}{0.47} & \multirow{2}{*}{\textbf{0.50}} & \multirow{2}{*}{0.46} & \multirow{2}{*}{0.49} & \multirow{2}{*}{0.00} \\
         &  &  &  &  &  &  &  &  \\
         & \multirow{2}{*}{\shortstack{2-D \\ 16 Demo}} & \multirow{2}{*}{-} & \multirow{2}{*}{0.31} & \multirow{2}{*}{0.50} & \multirow{2}{*}{0.50} & \multirow{2}{*}{0.51} & \multirow{2}{*}{\textbf{0.52}} & \multirow{2}{*}{0.00} \\
         &  &  &  &  &  &  &  &  \\
         \cline{2-9}
         & Antonyms & 0.47&	0.78&		0.59&	\textbf{0.84}	&	0.79&	0.79	&0.04 \\
         & Capital & 0.59&	0.86&		0.77&	\textbf{0.89}	&	0.83&	0.86&	0.14 \\
         & Language & 0.42&	0.75	&	0.33	&0.72	&	0.79&	\textbf{0.83}&	0.04 \\
         & Profession & 0.71&	0.46&		0.34	&\textbf{0.55}&	0.52&	0.53&	0.07 \\
         & Religion & 0.74&	0.88	&	0.64	&0.84	&0.86&	\textbf{0.90}&	0.00 \\
         
         \bottomrule 
    \end{tabular}
\end{table*}

\end{document}